\documentclass{article}
\usepackage{spconf,amsmath,epsfig}
\usepackage{amsfonts}
\usepackage{amssymb}
\usepackage{hyperref}
\usepackage[font=small,labelfont=bf]{caption}

\usepackage{breqn}
\newcommand{\CRediT}{\textsuperscript{\small CRediT}}

\title{SEMI-SUPERVISED LEARNING FOR HYPERSPECTRAL IMAGES BY NON PARAMETRICALLY PREDICTING VIEW ASSIGNMENT\CRediT}

\name{Shivam Pande\textsuperscript{1}, Nassim Ait Ali Braham\textsuperscript{2}, Yi Wang\textsuperscript{2}, Conrad M Albrecht\textsuperscript{2}, Biplab Banerjee\textsuperscript{1}, Xiao Xiang Zhu\textsuperscript{3}
\thanks{Corresponding author: shivam\_pande@iitb.ac.in}
\thanks{\textsuperscript{1}Indian Institute of Technology Bombay, Mumbai, India}
\thanks{\textsuperscript{2}German Aerospace Center (DLR), Oberpfaffenhofen, Germany} \thanks{\textsuperscript{3}Technical University of Munich, Munich, Germany}
\thanks{~\textit{Contributor Role Taxonomy} statement, \url{https://credit.niso.org} --- \textbf{Shivam Pande}: Data curation, Software, Visualization, Writing--original draft; \textbf{Conrad M Albrecht}: Conceptualization, Methodology, Resources, Writing--review \& editing, Supervision; \textbf{Nassim Ait Ali Braham}: Methodology; \textbf{Yi Wang}: Methodology; \textbf{Biplab Banerjee}: Supervision; \textbf{Xiao Xiang Zhu}: Funding acquisition}}
\address{}

\begin{document}

\maketitle

\begin{abstract}
Hyperspectral image (HSI) classification is gaining a lot of momentum in present time because of high inherent spectral information within the images. However, these images suffer from the problem of curse of dimensionality and usually require a large number samples for tasks such as classification, especially in supervised setting. Recently, to effectively train the deep learning models with minimal labelled samples, the unlabeled samples are also being leveraged in self-supervised and semi-supervised setting. In this work, we leverage the idea of semi-supervised learning to assist the discriminative self-supervised pretraining of the models. The proposed method takes different augmented views of the unlabeled samples as input and assigns them the same pseudo-label corresponding to the labelled sample from the downstream task. We train our model on two HSI datasets, namely Houston dataset (from data fusion contest, 2013) and Pavia university dataset, and show that the proposed approach performs better than self-supervised approach and supervised training. 
\end{abstract}
\section{Introduction}
\label{sec:intro}

With the innovation in the sensing technologies, we are continuously witnessing the increase in the amount of data being collected in the remote sensing domain. The domain of hyperspectral imaging in particular has been under significant progress with images being generated of high spectral and spatial resolution. However, the acquisition of humongous data leads to the challenges of data processing. With the current technology, most of the data processing task, specifically on a larger scale, are religiously handled using algorithms from deep learning (DL). Even though DL techniques have shown to perform brilliantly with the big data, most of the applications, specifically those from supervised learning require large number of annotated samples. Since data annotations depend on manual interference, it becomes necessary to come up with the techniques that rely on limited samples for a decent performance. Self-supervised and semi-supervised learning have shown to be efficient learning paradigms for such limited label learning, where the models are trained by leveraging the information inherent in the unlabeled data. 
\begin{figure*}[ht!]
  \centering
  \centerline{\includegraphics[width=17cm]{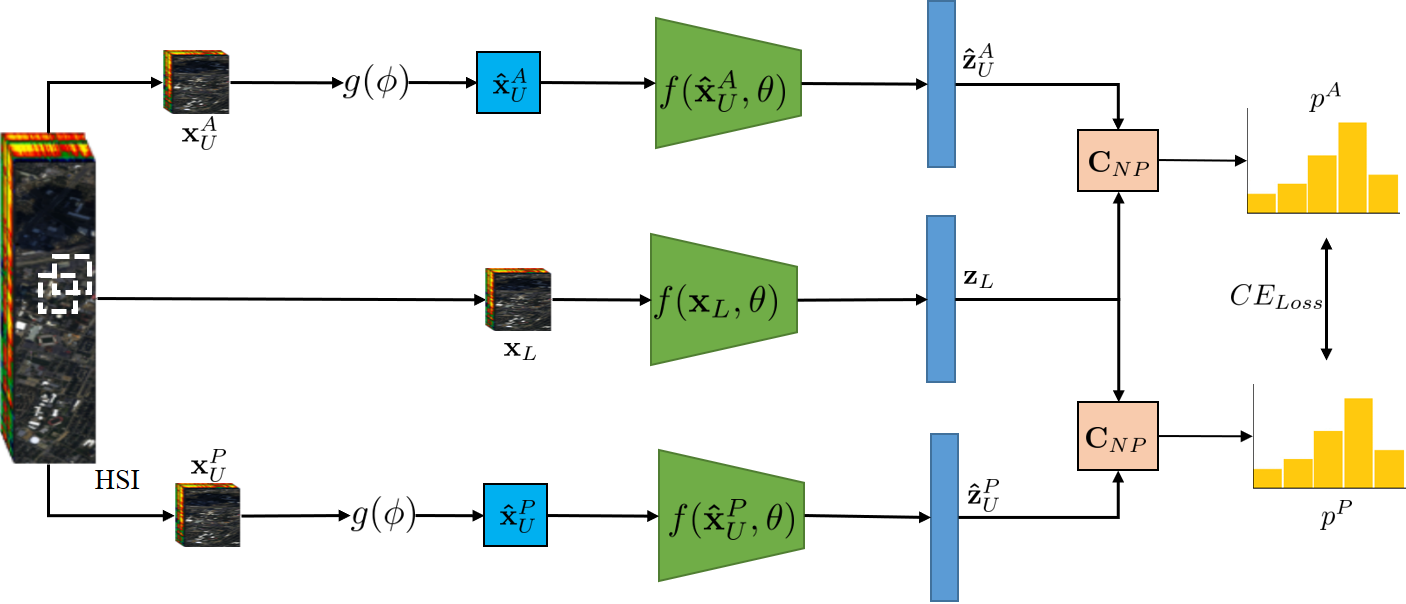}}
\caption{The figure shows the schematic of the PAWS model for semi-supervised pretraining. From the HSI, we obtain the unlabelled overlapping patches for anchor and positive sets. These patches pass through the augmentation function $g(\phi)$ to get the augmented representations. The augmented anchor and positive samples are sent through the encoder $f(\theta)$ to obtain the embedding $\hat{z}_U^A$ and $\hat{z}_U^P$. Similarly, for the labelled sample, the embedding $z_L$ is obtained. $\hat{z}_U^A$ and $\hat{z}_U^P$ are compared with $z_L$ using a non-parametric classifier $C_{NP}$. The pseudo-label probabilities obtained after classification are then compared using the cross-entropy loss on which the model is trained.}\medskip
  \vspace{-0.5cm}
\label{fig:paws}
\end{figure*}
Recently, several works pertaining to self-supervised learning have been studied with regards to hyperspectral image classification. One of the pioneer works in the self-supervised learning domain is presented in \cite{mou2017unsupervised}, where the authors propose a generative SSL based approach for HSI classification. Here, the network is initially pretrained as an autoencoder to reconstruct the original 3D HSI patch. The encoder part of the pretrained model is then used for classification of HSI samples. \cite{braham2022self} proposed a method for self-supervised learning in HSI classification. Their method was inspired from Barlow Twins method of SSL, where the pre-training happens by minimising the distance between the cross-correlation matrix between the embedding of the two augmented views and the ideal correlation matrix of HSI samples. Similarly, \cite{li2022robyol} presented a work leveraging the idea of \textit{bootstrap your own latent (BYOL)} in HSI classification. Additionally, they introduced the idea of random occlusion for augmentation in multiple views. 

Most of the works mentioned above purely relied on the idea of self-supervised learning only. However, it was proposed in \cite{assran2021semi} that an additional guiding information could be provided for pretraining by leveraging the training samples that would be anyways available in the groundtruth. Their work focussed on minimising a consistency loss between the pseudo-labels corresponding to the different augmented views, that are assigned with respect to the labelled samples from the groundtruth (where, they termed their approach as PAWS). Based on the aforementioned concept, we take an opportunity to build on the idea of semi-supervised learning for HSI classification, in the current research work. Our contributions are as follows:
\begin{itemize}
    \item We extend the idea of PAWS in the domain of hyperspectral image classification, where we show its performance compared to supervised and self-supervised algorithms.
    \item We evaluate our approach on Houston 2013 dataset in different settings (with and without data augmentations) to showcase its performance. 
\end{itemize}

\section{Methodology}

Let $\mathcal{X} \in \mathbb{R}^{M \times N \times B}$ represent the hyperspectral image with $M$, $N$ and $B$ the number of rows, columns and channels respectively. $\mathbf{x}_U^{j} \in \mathbb{R}^{p \times p \times B}$ represent the unlabelled samples, while $\mathbf{x}^{i}_L \in \mathbb{R}^{p \times p \times B}$ represent the labelled samples, with the labels of the central pixel assigned to each sample. Here, $p$ represents the spatial dimension of the sample. Let $y^{i}_L \in \mathcal{Y}_L$ represent the labels associated with the training samples. Here, $i\in\{1,2,...N_L\}$ and $j\in\{1,2,...N_U\}$, where $N_L$ and $N_U$ is the number of labelled and unlabelled samples respectively. The training process is divided in two stages, namely \textit{pretraining} and \textit{classification}, that are discussed in the subsequent sections. The schematic of the proposed approach is presented in Fig. \ref{fig:paws}.

\subsection{Pretraining}

This section explains the semi-supervised pretraining of the proposed PAWS model. Let $(x^i)^A_U \in \mathbb{R}^{p \times p \times B}$ denote the anchor view and $(x^i)^P_U \in \mathbb{R}^{p \times p \times B}$ denote the positive view, (which spatially overlap). The two views are passed through a non-trainable probabilistic augmentation function $g(\phi)$ to obtain respective augmented views as $(\hat{x}^i)^A_U$ and $(\hat{x}^i)^P_U$. Here, $\phi$ is the probability vector to choose the respective augmentation. Let $f(\theta)$ denote the encoder that maps the given hyperspectral patch to the hidden representation ($f(\theta): \mathbb{R}^{p \times p \times B} \xrightarrow{} \mathbb{R}^d$). We have used two kinds of encoders for the experimets. One of them is derived from a WideResNet \cite{zagoruyko2016wide} architecture with input layer changed to accommodate hyperspectral data. The other encoder is a fully convolutional framework with the initial layer being a 3D convolution layer and the later layers followed by three depthwise separable convolution layers, thus resulting in the $d$ dimensional vectors $(\hat{z}^i)^A_U$ and $(\hat{z}^i)^P_U$. Simultaneously, for a labelled sample $(x^j)_L \in \mathbb{R}^{p \times p \times B}$ (acquired from labelled \textit{support set}), we get the representation $(z^j)_L$. 

The support representation $(z^i)_L$ is then compared with the anchor and positive representations $(\hat{x}^i)^A_U$ and $(\hat{x}^i)^P_U$ using a cosine similarity based non-parametric \textit{soft nearest neighbour (SNN)} classifier \cite{salakhutdinov2007learning} ($C_{NP}$), given as:
\begin{equation}
(\hat{p}^j)^A = C_{NP}({z^j_U, \mathbf{z}_L}) = \sigma_\tau((z^j)^A_U\mathbf{z}^T_L)\mathcal{Y}_L
\end{equation}
Similarly, $(\hat{p}^j)^P$ can be calculated as well. Here, $\sigma$ is the softmax activation and $\tau (>0)$ is the temperature. The calculated probabilities are then sharpened to avoid representation collapse and get confident predictions. The sharpening function is denoted as:
\begin{equation}
[\rho(\hat{p}^j)]^k = \frac{([p^j]_k)^{\frac{1}{T}}}{\sum_{t=1}^K([p^t]_k)^{\frac{1}{T}}} 
\end{equation}
Here $k \in \{1, 2, ..., K\}$, where $K$ is the number of classes. 
For training the model, we simultaneously consider two cross entropy loss terms between $(\hat{p}^j)^P$ and $(\hat{p}^j)^A$. Additionally, an auto-entropy based regularizer derived from the average sharpened predictions is included. This is to ensure that the average prediction is close to the the mean distribution. 
The final loss is written as:
\begin{dmath}
\mathcal{L}_{SSL} = \frac{1}{2n}\sum_{i = 1}^{n}(\mathcal{E}(\rho((\hat{p}^j)^A), (\hat{p}^j)^P) + \mathcal{E}(\rho((\hat{p}^j)^P), (\hat{p}^j)^A)) - \mathcal{E}(\bar{p}) 
\end{dmath}
Here, $\mathcal{E}$ denotes the entropy function, while $\bar{p}$ is the average probability of the predictions, given as:
\begin{equation}
\bar{p} = \frac{1}{2n}\sum_{i = 1}^{n}(\rho((\hat{p}^j)^P), \rho((\hat{p}^j)^A))
\end{equation}
In the cross entropy loss, while training, the gradients are not calculated with respect to the sharpened targets. 
\subsection{Classification}
\label{sec:classification}
\begin{figure}[ht!]
  \centering
  \centerline{\includegraphics[width=8.5cm]{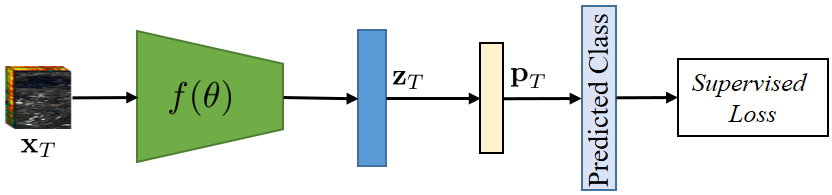}}
\caption{Schematic of linear classification and finetuning after PAWS based pretraining.}\medskip
  \vspace{-0.5cm}
\label{fig:ft}
\end{figure}

After pretraining, the model is evaluated for the task of landuse/land cover classification. We have applied three techniques as classification benchmarks:
\begin{enumerate}
\item \textbf{Linear layer}: A fully connected layer is appended over the existing model. The weights of the pretrained model are frozen, while only the weights of the final layer are modified using training samples. The model is then evaluated on the test samples. 
\item \textbf{Fine tuning}: It is also akin to the previous classifier, but instead of freezing the weights, all the weights are simultaneously modified using training samples. The model is then evaluated on the test samples. The schematic of linear classification and finetuning can be seen in Fig. \ref{fig:ft}.
\item \textbf{Soft nearest neighbor (SNN) classifier}: Here, the trained model simply acts as a feature extractor without further downstream optimization. Both training and test samples are passed through the pretrained model. The test samples are then compared with the training samples using a cosine similarity based classifier (such as classifier $C_{NP}$).
\end{enumerate}

\section{EXPERIMENTS AND RESULTS}
\label{sec:experimets}
\subsection{Datasets}
\label{ssec:datasets}
\begin{figure}[ht!]
  \centering
  \centerline{\includegraphics[width=8.5cm]{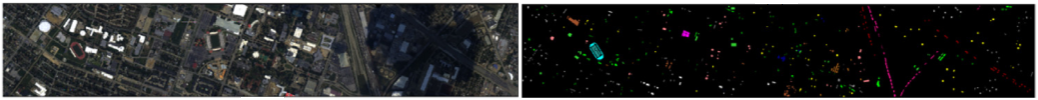}}
\caption{Houston 2013 dataset. False colour composite (Left). Groundtruth (Right)}\medskip
  \vspace{-0.5cm}
\label{fig:h13_data}
\end{figure}

We have considered the following two datasets for our experiments:

\noindent\textbf{Houston dataset}: This dataset is from data fusion contest, $2013$. The spatial extent of the HSI is $349$ $\times$ $1905$ and is composed of $144$ bands. The dataset consists of $2832$ training samples and $12197$ test samples, distributed over $15$ land use/land cover classes. For our experiments, we have selected $100$ samples per class from the training samples as support set (overall $1500$ patches). For the anchor and positive points, we have sampled out $71416$. The HSI and the corresponding groundtruth are presented in Fig. \ref{fig:h13_data} \cite{bose2021two}. 

\begin{figure}[ht!]
  \centering
  \centerline{\includegraphics[width=8.5cm]{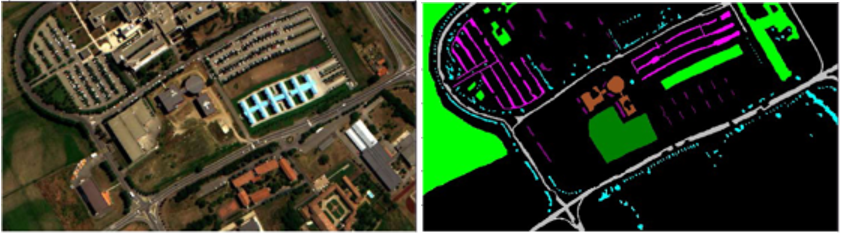}}
\caption{Pavia University dataset. False colour composite (Left). Groundtruth (Right)}\medskip
  \vspace{-0.5cm}
\label{fig:PU_data}
\end{figure}

\noindent\textbf{Pavia University dataset}: The hyperspectral image has the dimensions $610$ $\times$ $340$ $\times$ $103$, while the number of labelled samples is $42776$. For the anchor and positive points, we have sampled out $99000$ points, while the support set and test set are formed of $900$ ($100$ per class) and $41876$ points respectively, with the patch size of $9$. The HSI and the corresponding groundtruth are presented in Fig. \ref{fig:PU_data} \cite{braham2022self}. 

In both the datasets, the anchor and positive patches overlap by $67$\%.

\subsection{Training protocols}
\label{ssec:tp}
All the deep learning models use LARS optimizer \cite{you2019large} for pretraining stage. During classification, with linear layer as well as with fine tuning, stochastic gradient descent optimizer is used. In the experiments, the values of temperatures $\tau$ (in softmax) and $T$ (during sharpening) are fixed to $0.25$ and $0.10$ respectively. The number of epochs are fixed to 50 for the model pretraining, linear classification as well as finetuning. For every unlabeled sample, following augmentations have been applied:
\begin{enumerate}
\item \textbf{Spectral}: Random channel swapping, random channel dropping, channel intensity suppression, random channel 
averaging for a fixed number of channels
\item \textbf{Spatial}: Flipping (horizontal, vertical and mirror), random cropping, random rotation
\item \textbf{Spectral-spatial}: Pixel vector removal, noise introduction across channels and spatial dimension
\end{enumerate}

\subsection{Results and Discussions}
\label{sec:rnd}
The results are provided in Tables \ref{tab:results_h13} and \ref{tab:results_pavia} for Houston 13 and Pavia university datasets respectively. For Houston 13 dataset, a custom CNN is used, while the results of Pavia University dataset have been presented when trained on WideResNet. Column 1 indicates the model specifics. Column 2 denotes if augmentations have been applied. Column 3 represents the choice of classifier for downstream land cover classification, and corresponding overall accuracy values are listed in Column 4. It is visible that the proposed semi-supervised approach performs better than the supervised  approach, both in the linear layer setting as well as fine tuning setting. The method also performs better than one of the recently attempted self-supervised approach based on Barlow Twins loss. As baseline benchmark, SNN classification performance surpasses when the classifier is trained on features extracted from the pretrained model using our proposed approach. From Table \ref{tab:results_pavia}, it is also visible, that the accuracy on supervised classification is relatively much lower than the other models. This could be attributed to the large number of parameters in WideResNet in comparison to the custom CNN model. Best performances are highlighted in bold.
\begin{table}[htbp]
  \centering{\scriptsize
  \caption{Accuracy analysis for the proposed semi-supervised PAWS model on Houston 2013 dataset.}
    \begin{tabular}{|c|c|c|c|}
    \hline
    Encoder & Augmentations & Classifier & \multicolumn{1}{l|}{Accuracy} \\
    \hline
    \multicolumn{4}{|c|}{} \\
    \hline
    CNN   & No    & Supervised & 79.10\% \\
    Barlow Twins \cite{braham2022self}  & Yes   & Fine Tune & 77.14\% \\
    None  & No    & Linear & 64.21\% \\
    PAWS trained & Yes   & Linear & 80.92\% \\
    PAWS trained & Yes   & Fine Tune & \textbf{83.12}\% \\
    \hline
    \multicolumn{4}{|c|}{} \\
    \hline
    None  & No    & SNN   & 32.63\% \\
    Untrained & No    & SNN   & 34.28\% \\
    PAWS trained & Yes   & SNN   & \textbf{75.31}\% \\
    \hline
    \end{tabular}
  \label{tab:results_h13}}
\end{table}

\begin{table}[htbp]
  \centering{\scriptsize
  \caption{Accuracy analysis for the proposed semi-supervised PAWS model on Pavia University dataset.}
    \begin{tabular}{|c|c|c|c|}
    \hline
    Encoder & Augmentations & Classifier & \multicolumn{1}{l|}{Accuracy} \\
    \hline
    \multicolumn{4}{|c|}{} \\
    \hline
    CNN   & No    & Supervised & 21.38\% \\
    Barlow Twins \cite{braham2022self} & Yes   & Linear & 82.44\% \\
    None  & No    & Linear & 74.12\% \\
    PAWS trained & Yes   & Linear & \textbf{83.64}\% \\
    \hline
    \multicolumn{4}{|c|}{} \\
    \hline
    None  & No    & SNN   & 50.12\% \\
    Untrained & No    & SNN   & 46.33\% \\
    PAWS trained & Yes   & SNN   & \textbf{70.34}\% \\
    \hline
    \end{tabular}
    \label{tab:results_pavia}}
\end{table}

\section{Conclusions}
\label{sec:conclusion}
In this paper, we apply the method of semi-supervised learning (based on PAWS approach) for hyperspectral image classification. Our method leverages the labelled samples (available in the downstream task) to effectively guide the self-supervised pretraining process before the classification. We observe that the proposed model outperforms the supervised as well as self-supervised classification task, both in parametric and non-parametric settings. Motivated by our findings, we plan to extend our experiments to large-scale missions such as DLR’s EnMAP satellite launched in April 2021 and recently moved into the operational phase.

\section*{Acknowledgement}
This project is supported by the German Federal Ministry of Education and Research (BMBF) in the framework of the International Future AI lab ``AI4EO -- Artificial Intelligence for Earth Observation: Reasoning, Uncertainties, Ethics and Beyond" (grant number: 01DD20001). Additionally, this work is also supported by the Helmholtz Association through the Framework of \textit{HelmholtzAI}, grant ID: \texttt{ZT-I-PF-} \texttt{5-01} -- \textit{Local Unit Munich Unit @Aeronautics, Space and Transport (MASTr)}.  

\bibliographystyle{IEEEbib}
\footnotesize\bibliography{refs}

\begin{thebibliography}{1}

\bibitem{mou2017unsupervised}
Lichao Mou, Pedram Ghamisi, and Xiao~Xiang Zhu,
\newblock ``Unsupervised spectral--spatial feature learning via deep residual
  conv--deconv network for hyperspectral image classification,''
\newblock {\em IEEE Transactions on Geoscience and Remote Sensing}, vol. 56,
  no. 1, pp. 391--406, 2017.

\bibitem{braham2022self}
Nassim Ait~Ali Braham, Lichao Mou, Jocelyn Chanussot, Julien Mairal, and
  Xiao~Xiang Zhu,
\newblock ``Self supervised learning for few shot hyperspectral image
  classification,''
\newblock in {\em IGARSS 2022-2022 IEEE International Geoscience and Remote
  Sensing Symposium}. IEEE, 2022, pp. 267--270.

\bibitem{li2022robyol}
Jinhui Li, Xiaorun Li, Zeyu Cao, and Liaoying Zhao,
\newblock ``{ROBYOL}: Random-occlusion-based {BYOL} for hyperspectral image
  classification,''
\newblock {\em IEEE Geoscience and Remote Sensing Letters}, vol. 19, pp. 1--5,
  2022.

\bibitem{assran2021semi}
Mahmoud Assran, Mathilde Caron, Ishan Misra, Piotr Bojanowski, Armand Joulin,
  Nicolas Ballas, and Michael Rabbat,
\newblock ``Semi-supervised learning of visual features by non-parametrically
  predicting view assignments with support samples,''
\newblock in {\em Proceedings of the IEEE/CVF International Conference on
  Computer Vision}, 2021, pp. 8443--8452.

\bibitem{zagoruyko2016wide}
Sergey Zagoruyko and Nikos Komodakis,
\newblock ``Wide residual networks,''
\newblock {\em arXiv preprint arXiv:1605.07146}, 2016.

\bibitem{salakhutdinov2007learning}
Ruslan Salakhutdinov and Geoff Hinton,
\newblock ``Learning a nonlinear embedding by preserving class neighbourhood
  structure,''
\newblock in {\em Artificial intelligence and statistics}. PMLR, 2007, pp.
  412--419.

\bibitem{bose2021two}
Rupak Bose, Shivam Pande, and Biplab Banerjee,
\newblock ``Two headed dragons: {M}ultimodal fusion and cross modal
  transactions,''
\newblock in {\em 2021 IEEE International Conference on Image Processing
  (ICIP)}. IEEE, 2021, pp. 2893--2897.

\bibitem{you2019large}
Yang You, Jing Li, Sashank Reddi, Jonathan Hseu, Sanjiv Kumar, Srinadh
  Bhojanapalli, Xiaodan Song, James Demmel, Kurt Keutzer, and Cho-Jui Hsieh,
\newblock ``Large batch optimization for deep learning: Training {BERT} in 76
  minutes,''
\newblock {\em arXiv preprint arXiv:1904.00962}, 2019.

\end{thebibliography}

\end{document}